\documentclass[letterpaper]{article} 
\usepackage{aaai24}  
\usepackage{times}  
\usepackage{helvet}  
\usepackage{courier}  
\usepackage[hyphens]{url}  
\usepackage{graphicx} 
\usepackage{booktabs}
\usepackage{natbib}  
\usepackage{caption} 
\usepackage{algorithm}
\usepackage{algorithmic}

\usepackage{newfloat}
\usepackage{listings}
\DeclareCaptionStyle{ruled}{labelfont=normalfont,labelsep=colon,strut=off} 
\floatstyle{ruled}
\newfloat{listing}{tb}{lst}{}
\floatname{listing}{Listing}
\title{Beyond Thumbs Up/Down: Untangling Challenges of Fine-Grained Feedback for Text-to-Image Generation}

\author{
    Katherine M. Collins\textsuperscript{\thanks{Work conducted while at Google.}\rm 1},
    Najoung Kim\textsuperscript{\rm 2},
    Yonatan Bitton\textsuperscript{\rm 3},
    Verena Rieser\textsuperscript{\rm 2},
    Shayegan Omidshafiei$^*$\textsuperscript{\rm 4},Yushi Hu$^*$\textsuperscript{\rm 5},
    Sherol Chen\textsuperscript{\rm 2},
    Senjuti Dutta\textsuperscript{\rm 6},
    Minsuk Chang\textsuperscript{\rm 3},
    Kimin Lee$^*$\textsuperscript{\rm 7},
    Youwei Liang$^*$\textsuperscript{\rm 8},
    Georgina Evans\textsuperscript{\rm 2},
    Sahil Singla\textsuperscript{\rm 2},
    Gang Li\textsuperscript{\rm 3},
    Adrian Weller\textsuperscript{\rm 1, \rm 9},Junfeng He\textsuperscript{\rm 3},
    Deepak Ramachandran\textsuperscript{\rm 3}, Krishnamurthy Dj Dvijotham$^*$\textsuperscript{\rm 10}
}
\affiliations {
    \textsuperscript{\rm 1}University of Cambridge\\
    \textsuperscript{\rm 2}Google DeepMind\\
    \textsuperscript{\rm 3}Google Research\\
    \textsuperscript{\rm 4}Field AI\\
    \textsuperscript{\rm 5}University of Washington\\
    \textsuperscript{\rm 6}University of Tennessee\\
    \textsuperscript{\rm 7}Korean Advanced Institute of Science and Technology\\
    \textsuperscript{\rm 8}University of California San Diego\\
    \textsuperscript{\rm 9}The Alan Turing Institute\\
    \textsuperscript{\rm 10}ServiceNow\\
    kmc61@cam.ac.uk\\
    ramachandrand@google.com\\
    dvij@cs.washington.edu
}

\usepackage{bibentry}

\begin{document}

\maketitle

\begin{abstract}
Human feedback plays a critical role in learning and refining reward models for text-to-image generation, but the optimal form the feedback should take for learning an accurate reward function has not been conclusively established. This paper investigates the effectiveness of fine-grained feedback which captures nuanced distinctions in image quality and prompt-alignment, compared to traditional coarse-grained feedback (for example, thumbs up/down or ranking between a set of options). While fine-grained feedback holds promise, particularly for systems catering to diverse societal preferences, we show that demonstrating its superiority to coarse-grained feedback is not automatic. Through experiments on real and synthetic preference data, we surface the complexities of building effective models due to the interplay of model choice, feedback type, and the alignment between human judgment and computational interpretation. We identify key challenges in eliciting and utilizing fine-grained feedback, prompting a reassessment of its assumed benefits and practicality. Our findings -- e.g., that fine-grained feedback can lead to worse models for a fixed budget, in some settings; however, in controlled settings with known attributes, fine grained rewards can indeed be more helpful -- call for careful consideration of feedback attributes and potentially beckon novel modeling approaches to appropriately unlock the potential value of fine-grained feedback in-the-wild.
\end{abstract}

\begin{figure}[h]
\begin{center}
\includegraphics[width=0.45\textwidth]{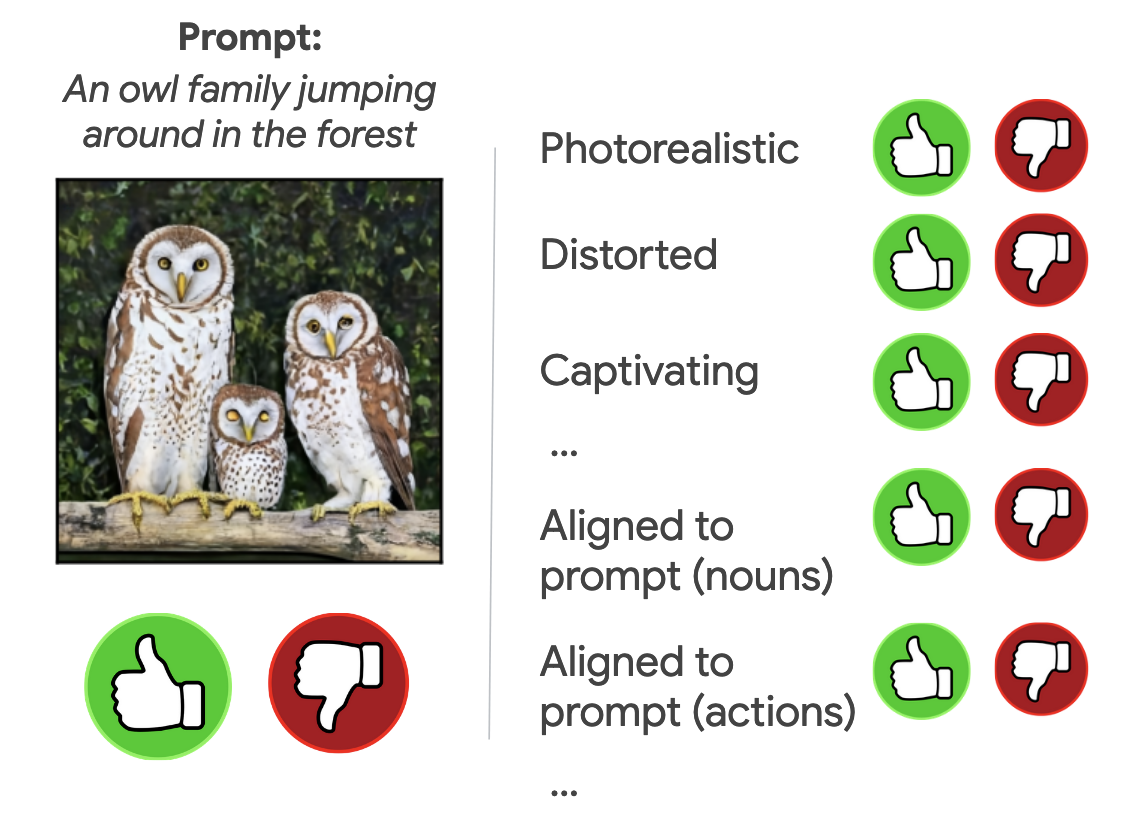}
\end{center}
\caption{Example text-image pair where granular feedback matters.}
\label{fig:intro}
\end{figure}

\begin{figure*}[h!]
\begin{center}
\includegraphics[width=0.8\linewidth]{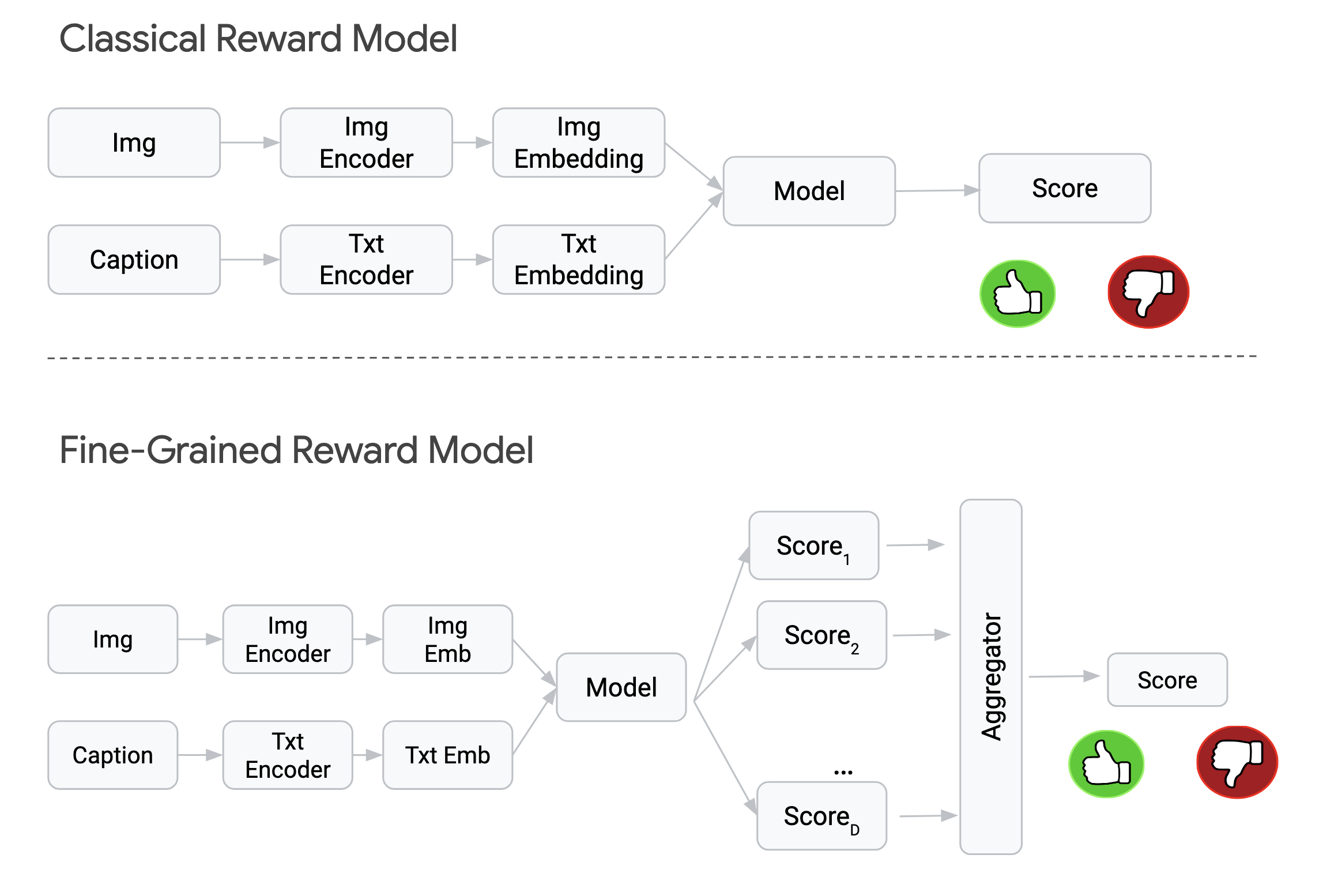}
\end{center}
\caption{Top: a typical coarse-grained feedback reward pipeline; bottom: proposed method for modeling fine-grained feedback.}
\label{fig:reward-pipeline}
\end{figure*}

\section{Introduction}
\label{intro}

Human feedback serves as a critical element in adapting large-scale generative models, particularly within the Reinforcement Learning with Human Feedback (RLHF) paradigm \citep{casper2023open,christiano2017deep, ouyang2022training}. However, conventional methods often rely on coarse-grained feedback, such as a single binary preference or Likert-scale rating, which may not adequately capture the nuances of quality and prompt-alignment in complex domains like text-to-image generation. A generated image may be highly visually appealing while deviating from the prompt, or conversely, align with the prompt and target visuals in some but not all desired ways, as illustrated in Figure \ref{fig:intro}. Recent research in text-to-text models (i.e. Large Language Models) suggests the potential of fine-grained feedback to address this challenge by enabling users to express their preferences with greater granularity, targeting specific features or interactions within the generated output \citep{wu2023fine, lee2023aligning, liang2023rich}. This granular feedback mechanism promises more precise control over model adaptation and behavior, which could lead to text-to-image models that exhibit enhanced responsiveness to diverse user needs and preferences. For example, a user may want more photorealistic owls for a presentation but less for a fun t-shirt~\citep{dutta2024understanding}.

This paper explores the complexities and trade-offs associated with utilizing fine-grained feedback for text-to-image model adaptation. We investigate its impact on model performance and explore strategies for integrating this nuanced information, taking into account the difficulty of elicitation. Our experiments reveal a complex interplay between model architecture, feedback type, and the alignment between human judgment and simulated AI preference judgments, ultimately influencing the effectiveness of fine-grained feedback.
In particular, we critically examine the hypothesis that reward models trained on fine-grained feedback exhibit superior performance in capturing human preferences within the text-to-image setting. Our empirical investigations, encompassing simulated and real human judgments alongside controlled scenarios, reveal surprisingly that while fine-grained feedback can provide an advantage under specific conditions, it does not consistently outperform coarse-grained feedback in the construction of effective reward models. In fact, coarse-grained feedback occasionally led to superior performance, highlighting the complexities of human preference modeling and the potential influence of architectural choices. However, when we do have complete knowledge of the attributes that may ``matter'' in preference judgements, as we explore through a de novo controlled experimental set-up, we do indeed illuminate the potential value of fine-grained feedback.

This work underscores the need for further exploration of alternative modeling paradigms capable of effectively harnessing the richness of fine-grained feedback while addressing the limitations of current approaches. Additionally, careful consideration of feedback attributes and task characteristics is crucial for maximizing the value and efficiency of incorporating human feedback into text-to-image model development.

Our key contributions include:
\begin{itemize}
\item  {\em Assessment framework}: We propose a framework utilizing rejection sampling as a proxy for large-scale adaptation, enabling efficient evaluation of fine-grained feedback utility.
    \item {\em Empirical case studies}: Our experiments demonstrate that the additional value of fine-grained feedback for training text-to-image reward models is highly conditional.

\item  {\em Open challenges}: We identify and discuss open challenges surrounding the construction and evaluation of reward models for text-to-image systems based on fine-grained feedback.

\end{itemize}

\section{Related Work}
\label{related-work}

The rise of generative AI systems and power of RLHF -- fueled by \textit{human feedback} -- has ignited further interest in gleaning insights from human data at scale to guide improved generation, e.g., to better align images to text prompts \citep{yarom2023you, hu2023tifa, kirstain2023pick, xu2023imagereward} or to personalize image-generation models \citep{vonrutte2023fabric, fan2023dpok}. Improved generative models offer immense potential to transform human productivity and creativity -- provided they adequately meet the diverse needs and preferences of users. It is not yet clear what kind of feedback is best to elicit to improve their output -- nor how best to incorporate such feedback when elicited. Works have taken steps to explore human feedback in text-to-image contexts ~\citep{liang2023rich, lee2023aligning} as well as at evaluation-time ~\citep{lee2024holistic}. Yet, the sheer computational scale of such generative models renders them challenging to systematically explore design choices around feedback elicitation and incorporation. Here, we take steps to address this gap by offering further empirical glimpses into the nuanced value of fine-grained feedback in generative AI applications.

Several prior works have investigated the interplay between richer forms of human feedback and model performance. Notably, recent research in the text-to-text domain has explored the potential of fine-grained feedback \citep{wu2023fine, lee2023aligning, liang2023rich, ouyang2022training}. In these studies, ``fine-grained'' refers to feedback that goes beyond simple binary judgments or single-score ratings, allowing users to target specific aspects of the output, such as factual accuracy, logical coherence, or stylistic elements. This granular feedback enables more precise control over model adaptation and behavior, leading to outputs that better align with diverse user preferences.
In addition to granularity, richer feedback can also encompass representations of disagreement and uncertainty in human labels. For instance, researchers have explored the value of eliciting and learning with traces of human uncertainty in the form of soft labels \citep{peterson2019human, Uma_Fornaciari_Hovy_Paun_Plank_Poesio_2020, collinsSoftLabel2022, collins2023humanmixup, sucholutsky2023informativeness}.

These studies have primarily focused on training classifiers for traditional machine learning tasks like image recognition, where human feedback is typically collected before training. However, practical applications often necessitate the ability to provide feedback on already trained models, potentially due to legislative requirements (e.g., Article 13 in the EU AI Act~\citep{EUAIAct}) or evolving user needs. Concept Bottleneck Models (CBMs), which map inputs to higher-level attributes before regressing a final target, offer a potential solution by allowing humans to provide feedback on a model's intermediate outputs \citep{koh2020concept}, which we explore in the remainder of our work for modeling fine-grained feedback. 

\section{Problem Setting} 
\label{problem-setting}

We first provide a primer on the task of learning a reward model for adapting a generative model; we then introduce our model. 

\subsubsection*{Reward Modeling}

Our goal is to learn a reward function $R_{\theta}: \mathcal{X} \rightarrow \mathcal{S}$ that takes in a set of features $x \in \mathcal{X}$ and produces a scalar score $s \in \mathcal{R}$ indicating the quality of $x$\footnote{The reward models we consider in this work produce pointwise, rather than pairwise, quality scores for each input.}. Many popular reward learning-based frameworks exist that adapt pre-trained language~\citep{ouyang2022training} or diffusion-based models~\citep{fan2024reinforcement, black2023training, dvijotham2023algorithms} to generate outputs with high reward scores. 

How do we build a good $R_{\theta}$? A popular approach is to learn the parameters of the reward model ($\theta$) from human feedback. That is, we curate a bank of examples D = \{$(x_1, s_1)$, $(x_2, s_2)$, ... $(x_N, s_N)$\} where $s$ are the result of human annotations. We can then update $\theta$ on any standard loss function $\mathcal{L}$ to improve our mapping from features to the score. Ideally, this produces a reward model $R_{\theta}$ that matches human preferences over ``good'' $x$.

\subsubsection*{Fine-Grained Feedback}
But what makes an $x$ ``good''? \citeauthor{casper2023open} raise several crucial open questions for learning reward models from human feedback, e.g., heterogeneity across humans and the challenge of going beyond single aggregate preferences. This task becomes all the more challenging when we consider generative models producing complex and highly structured outputs such as text-to-image models. For example, an image might be visually compelling and highly creative, but not necessarily aligned to either the main intention or minor attributes of the input prompt (or vice versa). Moreover, there may be multiple factors that together determine  the quality or aesthetic value of the image (e.g., it does not have artifacts, uses an appealing color palette  etc).

We consider the setting of eliciting \emph{fine-grained feedback} where human annotators are asked to provide  a set of $M$ scores $s_{i} = (s^{1}_{i}, s^{2}_{i}, ..., s^{M}_{i})$ for each example $x_i$ with $i \in {1,\ldots, N}$. The individual scores are scalar values representing the degree to which the prompt-image pair satisfies some particular aspect of quality (e.g. photorealism).  Our goal then is to learn a ``good'' (see Section \ref{sec:fine-grained}) $R_\theta$ from such feedback. We refer to reward models learned from more than one feedback as ``fine-grained reward models'' compared to those trained on a single aggregate score, $\tilde{s}$ (``coarse-reward models'') and discuss different ways to concretely operationalize ``good'' in our experiments.

\subsubsection*{Costs}

While there seems to be intuitive value to collecting finer-grained feedback over more attributes, the elicitation of such feedback necessarily incurs some additional time (or other resource) cost, over the cost of collecting coarse-grained feedback. We consider the setting where each $j = 1 \ldots M$ dimension of feedback is associated with some elicitation cost $c_{j} > 0$. For our computational experiments, we assume each form of feedback has equivalent procurement costs. We discuss deviations from this assumption in Section \ref{rejection-sampling}. 

\section{Reward Models from Fine-Grained Feedback} 
\label{sec:fine-grained}
What kind of model structure empowers us to effectively learn from such rich feedback? We consider a two-stage structure which first predicts each fine-grained attributes and then aggregates the scores, similar to \citep{wu2023fine}. This structure parallels a Concept Bottleneck Model \citep{koh2020concept} (see above), wherein our concepts our fine-grained attributes (is the image malformed? is the image blurry? is the image aligned to the text for verbs?). Rather than learn a mapping directly from $R_\theta: x \rightarrow \tilde{s}$, we learn two sets of parameters: $f_\phi: x \rightarrow s^{1}, ... s^{M}$ and $g_\psi: s^{1}, ... s^{M} \rightarrow \tilde{s}$. The functions then compose to produce a single aggregate score for a given input $g_\psi(f_\phi(x)) = \tilde{s}$ with the added benefit that we can \textit{inspect} the fine-grained attributes predicted rendering our model more interpretable.  

\subsubsection*{Our Model}

There are many functional forms that our two-stage modeling pipeline can take on. Here, we let $f$ be a multi-headed multi-layer perception (MLP). Following popular practice in the CBM literature, $g$ is a simple linear aggregator \citep{koh2020concept, cem22, margeloiu2021concept, collins2023human}. We consider the sequential CBM setting \citep{koh2020concept}, learning $\phi$ then $\psi$ separately. 

\subsubsection*{Embeddings}

Additionally, in the text-to-image setting, we need some way of providing the multimodal stimulus (text and image) as input to our model. Following standard practice, we use learned embeddings of the image pixels and text tokens respectively. We assume that the feature extractors which produce the embeddings are fixed (see Appendix\footnote{The full Appendix can be found in the arXiv version of this paper.}). This approach has a benefit in common real-world scenarios where practitioners only have access to features in black-box fashion. 

\section{Experiment Outline} 
\label{exps}

It is natural to expect that more informative supervision (finer-grained feedback) will be better for RLHF scenarios.  In particular, we hypothesize that fine-grained feedback will be valuable when training reward models in regimes with few examples, following the literature around informativeness of label supervision~\citep{sucholutsky2023informativeness}.
We posit based on previous results in text-to-text generation e.g. \citep{wu2023fine} that reward models trained from fine-grained feedback will be able to better capture preference judgments than a model trained on coarse preference judgments alone.

To address our hypothesis, we design a series of computational experiments with feedback of varying levels of granularity and dataset sizes. We consider two classes of fine-grained feedback important for measuring the quality of a text-to-image generation following \cite{lee2024holistic}: 
\begin{enumerate}
    \item Image quality: whether the image itself meets a desired criteria (e.g., photorealistic, not malformed). 
    \item Text-image alignment: whether the image is aligned to the text according to a particular semantic category (e.g., attributes from the prompt are captured in the image). 
\end{enumerate}

Recall in our reward model that we employ a two-stage pipeline: first predicting fine-grained feedback, then predicting aggregate targets. As such, like in CBMs, we need two sets of labels (over fine-grained attributes and aggregate targets). In this work, the additional fine-grained feedback signals are  inferred  by querying the state-of-the-art PaLI model \citep{chen2022pali} (see Section \ref{human-prefs-exps}) --- we leave the expansion of collecting and incorporating granular feedback from human annotators for future work. In all experiments, the coarse- and fine-grained models are trained on the same final targets --- real human preference judgments in Experiment 1 and a synthetic target permitting more controlled exploration in Experiment 2.

We consider two experimental settings. In the first, we explore the impact of fine-grained feedback for capturing real human preference judgments. In light of our negative result on the utility of learning CBM-based reward models from fine-grained feedback, we design a second, controlled domain to further disentangle the source of the poor performance.

\section{Experiment 1: Predicting Human Preference Judgments from Fine-Grained Feedback}
\label{human-prefs-exps}

\begin{figure*}[h!]
\begin{center}
\includegraphics[width=\textwidth]{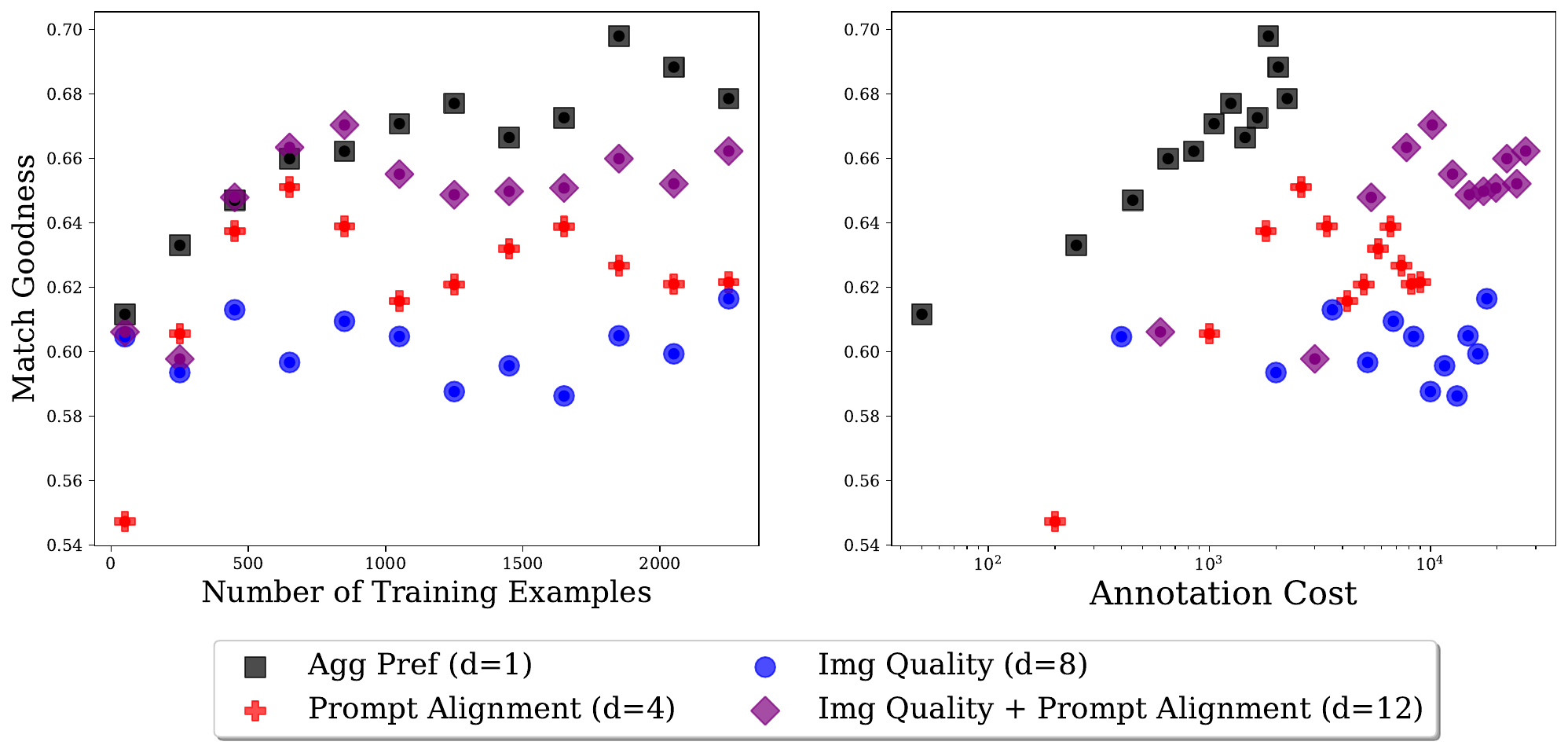}
\end{center}
\caption{Comparing reward models trained on coarse feedback (i.e., direct human preference judgments; black) against CBM-based models learned from fine-grained feedback. Reward models are differentiated by whether they were trained on granular feedback only about image quality (blue), image-text alignment (red), or both (purple). Left: Each point represents a reward model trained on $N$ image-prompt examples (x axis); ROC-AUC of the binary reward against held-out human preference judgments is presented on the y axis. Higher is better. Right: The same reward models, where the x axis (presented on a log scale) depicts estimated annotation cost, if each attribute is assumed to be equally costly to procure.}
\label{fig:agg-granular}
\end{figure*}

In this section, we consider the task of predicting real human preference judgements. We first overview our experimental set-up before presenting our results. We close with a discussion of what may underlie our observations that fine-grained feedback may not be preferable in this setting.
\subsection{Experimental Details}

\subsubsection*{Data} We use the approximately 5k images from ~\citep{dutta2024understanding}. Images were generated from DALL-E ~\citep{ramesh2021zero} and Stable Diffusion~\citep{rombach2022high} for over 1.3k text prompts from PartiPrompts ~\citep{yu2022scaling}. Each image has been annotated by nine humans, where images are scored rated on a scale of 1-4 for how ``satisfied'' the viewer is with the image for the intended prompt, with a particular motivational context in mind (i.e., participants were asked to rate how good an image-prompt pair is for a particular motivational context: for use as a phone background, graphic t-shirt, or presentation slide-deck). Here, we average all context-conditioned scores to form a single preference (``goodness'') score per image-prompt pair. This score forms our ``coarse''-grained preference of interest. We conduct a 50/25/25 train/val/test split at the level of the prompts (as there are four images per prompt). Additional details on data processing are included in the Appendix. 

\subsubsection*{Fine-Grained Attributes} The generative output of text-to-image models can be scored along both fidelity of the image to the prompt (text-image alignment~\citep{yarom2023you}) and image quality. We consider fine-grained attributes along each type: image quality and prompt alignment. 
For each image, we elicit granular quality attribute scores by querying PaLI \citep{chen2022pali} as discussed above. PaLI is a multimodal model which takes as input image and text and produces a text response; we can therefore use the model to simulate crowdsourced responses to text-image pairs by asking for a numerical score along some dimension. In particular, we train our model on eight image quality attributes (whether the image is \textit{distorted, photorealistic, bright, captivating, chaotic, visually compelling, disturbing, or funny}). We select these attributes to span a range of valence and relevance. Additional details on granular queries are included in the Appendix. 

To gather granular feedback for text-image alignment, we employ the $VQ^2$ framework from \citeauthor{yarom2023you}. $VQ^2$ evaluates image-text alignment by generating question-answer pairs from the input text. For example, the text ``A black apple and a green backpack'' could yield the question-answer pair ``What is the fruit in the picture? - Apple'' or ``What color is the backpack? - Green''.  These pairs are then assessed against the image using a Visual Question Answering model, which produces a ``Yes'' probability indicating the validity of each pair. The average of these ``Yes'' probabilities across all question-answer pairs constitutes the final $VQ^2$ score, reflecting the overall image-text alignment. We group $VQ^2$ questions  using a customized modular semantic parser into four attribute classes, whether the image matches the text in regards to: \textit{any actions/verbs mentioned in the prompt, any attributes/adjectives, any objects/nouns, or any relations} mentioned in the prompt. See Appendix for further details on our feedback extraction. 

\subsubsection*{Models} 

We compare a reward model trained directly on coarse feedback (the aggregate human scores from \citeauthor{dutta2024understanding}) against a suite of CBMs trained on varying amounts ($d$) and types (e.g., image-based or prompt-based) fine-grained attributes. Modeling details are included in the Appendix. Again, each attribute of fine-grained feedback acts like a ``concept'' predicted in the first stage of our two-stage model (Figure~\ref{fig:reward-pipeline}.

\subsubsection*{Evaluation} We evaluate models in two ways: 

\begin{itemize}
    \item Accuracy of the reward models, in terms of predicting the aggregated human ``goodness'' score for held-out examples from ~\citep{dutta2024understanding} (scored with ROC-AUC), and 
    \item Simulated adaptation of a downstream text-to-image generative model through rejection sampling, wherein we use our reward models to score generated examples, and check whether people agree with the relative rankings (i.e., that the stimulus rated higher by a fine-grained model indeed is preferred to a human over a stimulus rated highly by a coarse-grained model). We design and conduct a series of human evaluations here along aggregate and fine-grained dimensions. 
\end{itemize}

\subsection{Results}

\label{sec:exp1-results}
We compare a reward model trained directly on the coarse human preferences judgments against our CBM-based fine-grained models built from varying classes of fine-grained feedback. We train the suite of models over varying number of training examples and test on held-out image-prompt pairs. 

Interestingly, in Figure \ref{fig:agg-granular}, we find that training reward models simply on coarse-grained feedback is not only more cost-efficient for a given budget (i.e., if we assume equal costs for collecting labels for each additional fine-grained attribute), but yield better fits than the putatively more ``information-rich'' fine-grained reward models. We discuss the impact of deviations from the equal-cost assumption in Section \ref{rejection-sampling}. 

We do see that combining information about image quality with text-alignment boosts performance (see Figure \ref{fig:agg-granular}) over image quality information alone, but it is clear that the extra information in such attributes is not inducing higher match to held-out (in-distribution) preference judgments versus simply using the coarse-grained feedback. This raises the question that if such feedback has ``more information'' why are we seeing a performance drop compared to the coarse-grained model? We posit several hypotheses for why this may be the case. There may be:
 \begin{enumerate}
     \item Challenges stemming from the human data, e.g. our targets here are averages of human judgments originally produced for separate use cases (it may be better to model the full distribution), and by the motivation of our work, we may not want to try to match; 
     \item Misalignment between the synthetic feedback and real human judgment, in the values that we produce from PaLI and $VQ^2$ on the one hand, and the attributes that we collect from raters;
     \item Model expressivity: CBMs may not be adequately expressive to capture the nuances in the fine-grained feedback;
     \item Or, it could be that fine-grained feedback is fundamentally not useful here and provides no added value to the aggregate attributes.
 \end{enumerate}

\subsubsection{Evaluation Proxy for Adapting a Generative Model with Reward Models}
\label{rejection-sampling} 

We take a step to address the first point by running a fresh human evaluation. Recall, one of our goals for learning reward models from fine-grained feedback is to better tailor adaptation of downstream generative models and improve the quality of their output rather than just matching aggregate human preference judgments -- we made a case in our motivation that aggregate preference judgments can obfuscate important information. We simulate adapting a downstream generative model, and getting a sense of preference along the fine-grained attributes (focusing on the image quality attributes for simplicity). Due to computational costs, it is not always sensible to test out the quality of a gamut of reward models by adapting text-to-image models (see Section \ref{general-discussion}). Instead, we follow \citeauthor{lee2023aligning} in running \textit{rejection sampling} with our reward models as a proxy for adapting a generative model directly. That is, we draw samples from a generative model and use our reward models to \textit{score} the outputs; we then run head-to-head preference judgments over the generations favored by the respective reward models. Here, we consider two reward models: the model trained only on coarse judgments (i.e., the aggregate human preference judgments from the train set) and our model trained on fine-grained simulated attribute annotations\footnote{We consider our most ``fine-grained''-trained model with $d=12$ attributes}. Details on the sampled images and annotation procedures are included in the Appendix.

We find in Table \ref{tab:sxs-pref} (first row) that there is not a clear preference for the images that the reward model trained on coarse-grained feedback preferentially sample compared to the model trained on fine-grained feedback (i.e., annotators indicate that they prefer the text-image prompt rated more highly by the coarse-grained model than the fine-grained model for about 25\% of the examples we survey, and vice versa for fine- over coarse-). This finding suggests that the difference between the coarse- and fine-grained feedback trained models are not as strong as our in-distribution prediction-based evaluation (from Sec. \ref{sec:exp1-results}) make them appear as we move out-of-distribution to a new task: scoring generated images instead of judging (in-distribution) reward model accuracy. However, we observe high rates of uncertainty in the human judgements of which image is better along each attribute (annotators express that they are unsure which image they prefer for approximately 50\% of the samples). 

Such high levels of annotator uncertainty are exacerbated when we elicit judgments over individual dimensions (see Table \ref{tab:sxs-pref}). We observe strong signals only along the distorted and brightness dimensions. These findings suggest: 1) the preference for the results of the coarse-grained model are not consistent, and 2) eliciting preferences over fine-grained attributes may not be particularly meaningful nor informative. We might observe more interesting preference judgments along granular attributes with a different stimuli pool. Nonetheless, our results urge caution on the blind elicitation and incorporation of more granular feedback from annotators -- more is not always better (or at least not always informative). Further, we observe that, contrary to our simplifying assumptions, annotation times per dimension are \textit{not} uniform (see Table 
\ref{tab:sxs-time}), underscoring the importance of judiciously recognizing when to collect fine-grained attributes, and which to collect.

\begin{table}[h!]
\begin{tabular}{@{}llll@{}}
\toprule
\textbf{Feedback Type} & \textbf{Coarse} & \textbf{Fine-Grained} & \textbf{Unsure} \\ \midrule
Aggregate              &       25.6           &         24.9             &      49.5           \\ \midrule

Distorted              &       36.9           &     31.8                 &      31.3           \\
Bright                 &       30.2           &     26.1                 &      43.6           \\
Captivating            &       18.4           &     19.1                 &      62.5           \\
Photorealistic         &       31.1           &     31.4                 &      37.5           \\
Chaotic                &       13.7           &     12.4                 &      73.9           \\
Visually compelling    &       20.6           &     15.8                 &      63.6           \\
Disturbing             &       8.2           &      8.6                  &      83.2           \\
Funny                  &       0.5            &     0.9                  &      98.6           \\ \bottomrule
\end{tabular}
\caption{We compare pairwise general preferences (top row) as well as preference along particular granular attributes (rows below the line). Scores depict \% of images where coarse- vs. fine-grained were preferred (or people were uncertain), where \% depict the votes for each preferred image over the total number of votes.} 
\label{tab:sxs-pref}
\end{table}

\begin{table}[]

\centering
\begin{tabular}{@{}ll@{}}
\toprule
\textbf{Feedback Type} & \textbf{Time (s)} \\ \midrule
Aggregate              &       52.7        \\ \midrule
Distorted              &       56.1           \\
Bright                 &       18.4           \\
Captivating            &       20.2          \\
Photorealistic         &       19.4       \\
Chaotic                &       24.1           \\
Visually compelling    &       16.2          \\
Disturbing             &       19.2           \\
Funny                  &       12.8       \\ \bottomrule
\end{tabular}
\caption{Average annotator answer time (in seconds) for each annotation task.}
\label{tab:sxs-time}
\end{table}

\section{Experiment 2: Controlled Granular Image Quality Assessments}
\label{dectree-exps}

As mentioned, there could be a variety of reasons that we observe a null result for the utility of fine-grained feedback in Experiment 1. Crucially, we do not know what attributes people are considering when they are making their preference judgments. Perhaps if we elicited the correct attributes, we would be able to learn better reward models? As such, we are motivated to create a more controlled experimental setup where we \textit{do} know the attributes that are being considered in the final preference judgment.  

To address this gap, we design a second domain wherein we have complete knowledge over the attributes that inform the target preference. Unlike in Experiment 1, the target here is completely synthetic -- we build a decision tree over the fine-grained attributes (obtained from simulated AI feedback) that exactly determines the quality of an attribute\footnote{As in Experiment 1, we predict a single quality score per point, not a preference rating.}. This enables us direct control, but again, necessitates cautious interpretation as it side-steps the question (which may drive the null result in Experiment 1) of whether our AI-based feedback is even aligned with the judgments humans -- or particular humans -- make.

For simplicity, we focus on the case of only image-dependent evaluation (i.e., just considering the image attributes along and not those of the prompt). Since the ground truth is exactly captured by a custom decision tree with the simulated AI  feedback attributes as leaves; adequately capturing and modeling each dimension of granular feedback (i.e., the leaves), should be sufficient to learn reward models that accurately predict the target quality score. We emphasize that we construct this experiment to explore the impact of feedback granularity where we have direct access to the target (and know that it is constructed from multiple attributes); this set of experiments are \textit{not} indicative of which attributes matter for human aesthetic judgements.

\subsection{Experimental Setup} 
\label{experimental-setup}

\subsubsection*{Data and Evaluation} We consider the same images from ~\citeauthor{dutta2024understanding} as in Experiment 1. For simplicity, however, we consider \textit{only the images}; our experiments in this Experiment do not depend on the prompt.

\subsubsection*{Controlled Target} We design a controlled and fully intepretable target preference score formed from simulated attributes that enables us to more precisely understand the impact of granular feedback than trying to capture potentially nebulous real human preference scores. Specifically, we design a decision tree which takes in an image and at each node assesses a particular attribute; specifically, it checks whether it is photorealistic, then visually compelling, then chaotic. The output is a binary score which we take as representing whether an image is ``good'' or ``bad''.  

We evaluate reward models with ROC-AUC on held-out decision tree scored examples. We reiterate that the decision tree is intended to serve as a controlled target where we know which attributes underlie the final preference score; we do not claim this decision tree models human preference judgments nor is generalizable in all contexts.

\begin{figure}[h]
\begin{center}
\includegraphics[width=0.4\textwidth]{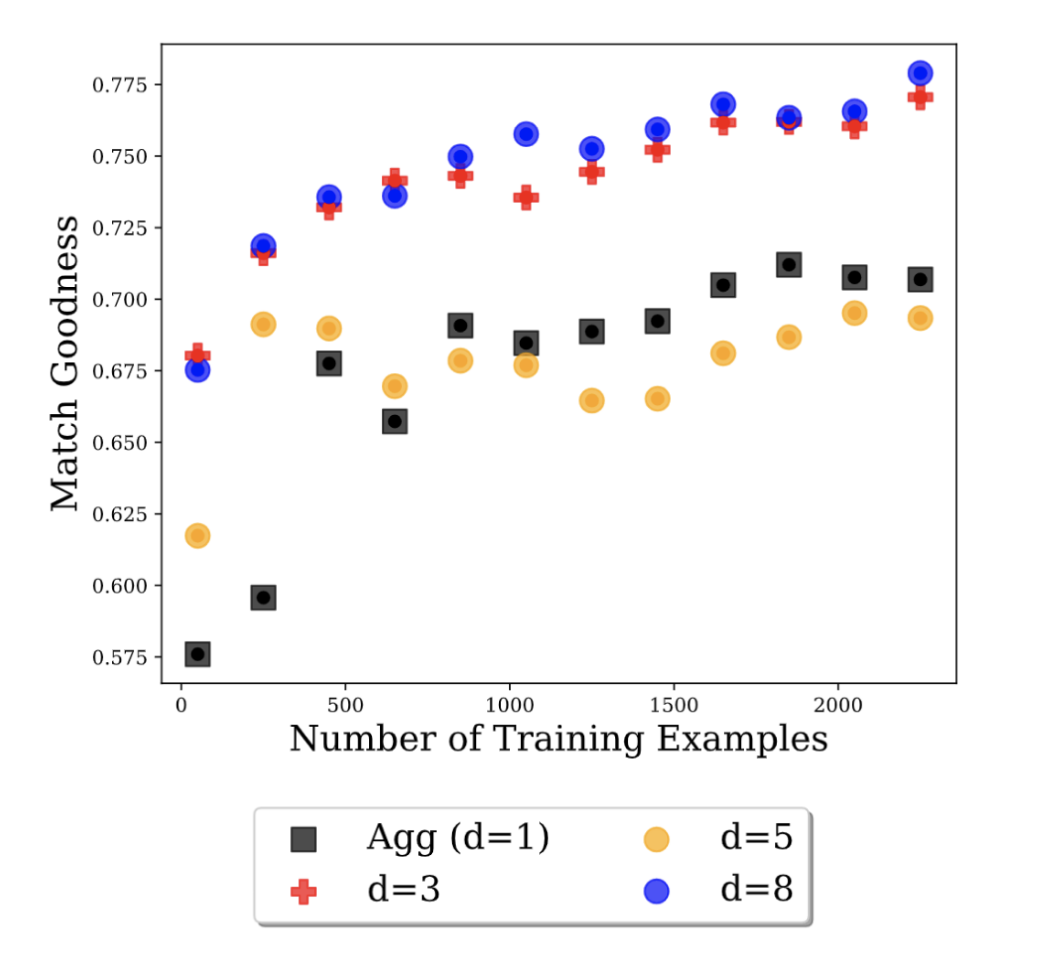}
\end{center}

\caption{Comparing reward models trained on varying levels of granularity. As in Figure \ref{fig:agg-granular}, each point represents a reward model trained on $N$ images. Models are scored according to the contrived decision tree on held-out examples. We compare a model trained directly on the single scalar decision tree scores (black) against a suite CBM-based fine-grained models trained on: 1) the same three attributes which make up the decision tree (red), 2) the same three attributes as the decision tree along with the remainder of the full set of image attributes under consideration (blue), and 3) only attributes \textit{not} included in the decision tree (orange).} 
\label{fig:dectree-goodness}
\end{figure}

 \subsubsection*{Models} We employ the same model architectures and training procedures as in our Experiment 1 experiments, with the exception that we only feed the image embeddings as input (further explorations of joint text-image modeling are important next steps). We train a coarse-grained model on the final output of the controlled decision tree, and compare this model against fine-grained CBMs which have access to varying numbers of image attributes (which may or may not include the attributes used to form the decision tree). 
 
\subsection{Results}

We start by considering the setting where our fine-grained CBM has access to the \textit{same} attributes as in the target decision tree; i.e., we compare a model trained directly on only the coarse score from the decision tree versus a CBM trained over the same attributes that make up the decision tree. Here, by design of our controlled decision tree, we ought to achieve the same or better performance to the coarse-grained setting; indeed, we do find in Figure \ref{fig:dectree-goodness} that we can achieve better performance by training on fine-grained feedback. Of note, adding more attributes beyond those in the ``true'' decision tree have little positive impact on reward quality, and may not be economical to elicit. While these data indicate that fine-grained feedback can be used to learn better reward models than those from aggregate preference judgements alone, one may ask why there is any gap between our learned reward models and the maximal achievable fit (which should be 100\% as the target is formed from in-distribution feedback). The gap suggests that our reward model is not as strong as it could be, possibly stemming from our embeddings (as discussed in the Open Questions). However, qualitative inspection of the reward models in the Appendix reveals that the fine-grained attribute models do reflect meaningful differences along the attributes, for example, distinguishing between images that are photorealistic or not.

Barring model architecture selection -- we make a crucial assumption -- that we know the true attributes. What if we do not have access to the attributes that form the decision tree? To begin to explore this question, we consider the setting where there is an \textit{attribute mismatch} (yellow points in Fig \ref{fig:dectree-goodness}). Here, we see a dramatic drop in performance. These data point to the importance of incorporating the right attributes if building a fine-grained reward model; here, we simulated and precisely controlled what attributes matter (by design of our decision tree). In practice, we may not know which attributes are ``right'' to elicit: the precise situation we found ourselves in for Experiment 1, potentially underlying our null result. One could envision selecting attributes that are most correlated with the target (see Appendix), but this requires having access to target annotations. Determining what attributes to elicit is a key open challenge, as we discuss next.


\section{Open Challenges} 
\label{general-discussion}

Nicely, we indeed find in Experiment 2 that fine-grained feedback can be useful to model if we know the attributes. But crucially, how do we actually \textit{find} these attributes? Our work urges further study of what attributes \textit{humans} consider when making preference judgments and what is \textit{economical} to elicit. Our work exposes key challenges that arise in the study of the impact of choice of feedback on reward models for adapting generative models.  

\subsection{Which Attributes to Elicit?}

Our work points to a key challenge for practitioners interested in collecting fine-grained feedback: what attributes should you elicit? In Experiment 2, we demonstrate that reward model performance may suffer if the elicited attributes do not match those that form the target preferences. How can we know what attributes we should elicit?  Such a question grows more challenging when we consider individual differences. Different attributes may matter to different people and depend on context \citep{gordon2022jury, kirk2024prism, dutta2024understanding}. We do not want reward models to collapse to a monoculture \citep{kleinberg2021algorithmic, bommasani2022picking}, but also ought to be mindful of the risks of personalization~\citep{kirk2024risks}. 

Additionally, it is not enough just to have the ``right'' attributes. In practice, elicitation needs to balance informativity and cost. We already see that attributes may take different amounts of time to annotate in Experiment 1 (see Table 
\ref{tab:sxs-time}) and in Experiment 2, we demonstrate that, in simulation, adding attributes is not always valuable. Interdisciplinary works that straddle AI, cognitive science, and human-computer interaction are already exploring the impact of requiring humans to provide feedback on many attributes, noting that such a practice can overwhelm cognitive load and risk bringing \textit{more error} into downstream modeling \citep{sucholutsky2023informativeness, ramaswamy2022overlooked, barker2023selective}. Indeed, we do not want to waste annotations on attributes where users are highly unsure (though future work can explore the benefits from learning with uncertainty at feedback time~\cite{collins2023human}). And further, we already see in our human studies that annotators spend substantially different time annotating some attributes over others. Nicely, the CBM model class naturally supports the implementation and study of cost-aware acquisition strategies for human feedback ~\cite{chauhan2022interactive, sheth2022learning, shin2023closer, espinosa2024learning}. We see promise in the hybridization of elicitation development for such models and the determination of which attributes to elicit.  

\subsection{Reward Model Structure}

Yet, perhaps CBMs are not the best model structure for fine-grained feedback. Indeed, recent work has raised questions about the ability of CBM-based systems to effectively handle rich, soft-labeled feedback if not explicitly trained to do so \citeauthor{collins2023human}. This highlights the importance of considering the interplay between feedback type and model architecture when designing systems for human-in-the-loop adaptation. We reiterate that our work is a preliminary exploration of ways to learn reward models from fine-grained feedback. It is likely that alternate modeling choices induce different cost-benefit analyses on the value of learning from fine-grained feedback; we look forward to future works that explore such possibilities. For instance, several other approaches have been proposed e.g. ~\citep{rame2024rewarded, liang2023rich}. We see the design of model architectures which incorporate information efficiently from granular feedback, and can flexibly grow to handle new dimensions (e.g, if we learn that a new attribute actually matters more to annotators that we had not previously modeled), as ripe for future work. Moreover, the image and text embeddings we considered in our work were always fixed. It is possible that different choices of embedding, or even jointly learning embeddings, may improve performance and perhaps salvage the utility of a CBM-based architecture.

\subsection{Accessible, Efficient Evaluation}

However, rapidly evaluating such modeling choices in the context of assessing reward models is not easy. The massive computational overhead of actually training and adapting large-scale generative models poses a crucial practical challenge for researchers attempting to study what kind of feedback yields powerful reward models. In our work, we attempted to deal with these challenges in two ways: 1) computational experiments wherein we have direct access to the target, and 2) simulating the impact of adapting a generative model downstream through our rejection sampling paradigm. While we hope our experimental approach illuminates one potential workflow that other researchers can take, more work is needed to characterize how much of a gap there is between such proxy settings and at-scale generative model adaptation.

\subsection{Human vs. Model Feedback}

Computational overhead is not the only challenge: we are also limited by the elicitation of feedback itself. Eliciting information from humans can be expensive. Here, our granular feedback was derived from an AI system, not humans. A natural question is how well our simulated feedback here actually correlates to human judgments. It is possible that our null results in Experiment 1 stem from a mismatch between human and model judgments over the granular attributes, either or both along the image quality and text-image alignment dimensions. While there is a push to employ AI-generated feedback rather than humans for scalable generative evaluation ~\citep{wu2023llms, gilardi2023chatgpt}, it is essential to understand where such feedback may diverges from human expectations~\citep{collins2024buildingmachineslearnthink}.

\section{Conclusion}

In this work, we uncover at least one setting where fine-grained feedback may not help immediately, under particular caveats (model choice, embedding efficacy, fidelity of fine-grained feedback, choice of attributes, minimal fine-tuning). Our work urges practitioners to consider carefully, particularly under a fixed annotation budget, what kind of feedback is useful and efficient to collect. It may not always make sense to collect fine-grained feedback -- and even if it does, some attributes may be more valuable than others. We need more interdisciplinary studies to identify what attributes people are considering and how well they align with model-derived feedback, and which attributes are worth encouraging people to consider to inform preference judgments for adapting text-to-image models. We hope our work inspires further study of efficient and robust ways of interleaving human and machine computation to study and improve generative models in a way that reflects the nuance replete in the world in which such systems are being deployed.

\section*{Acknowledgements}
We thank Ravi Rajakumar, Ellie Pavlick, and Sunny Mak for helpful feedback on the data used in the work. We additionally thank Jordi Pont-Tuset, Isabelle Guyon, and Andreas Terzis for helpful conversations and Alena Butryna for tremendous help coordinating our participant studies. AW acknowledges  support  from  a  Turing  AI  Fellowship  under grant  EP/V025279/1, and the Leverhulme Trust via CFI.

\bibliography{main}

\newpage

\section{Appendix}

\subsection{Image and Text Embeddings}
\label{embedding-details} 

Image and text inputs are represented as dense embedding vectors. We use CLIP to extract embeddings for text captions \citep{radford2021learning}. Through preliminary experimentation, we found the frozen CLIP embeddings  has been shown to poorly capture aesthetic properties of images; as such, we opted to extract the output of the first layer of the LAOIN Aeshetics model \citep{schuhmann2022laion} as our image embeddings. Embeddings are concatenated across modalities for Experiment 1; only the image embeddings are fed as used as input for training and inference in Experiment 2. Image and text embeddings are frozen for all models. Future work could explore the impact of jointly fine-tuning the embeddings and predicting granular feedback.

\subsection{Reward Models}

As introduced in Section \ref{problem-setting}, we run a two-phased training procedure for fine-grained reward models. We first train a mapping from the input embeddings (described above) to individual attributes; this takes the form of a multi-headed MLP. We then learn a simple linear aggregator over the outputs of the multi-headed MLP. All stages leverage binary classifiers for a form of feedback (specifically, multi-class binary classifiers when we have multiple attributes); the input to the Stage 2 linear aggregator for all settings is the sigmoided logit from Stage 1. The coarse-grained baseline only involves stage one (we directly map from the input embeddings to the coarse score); i.e., the coarse-grained model is \textit{not} a CBM. 

We emphasize that alternate ways of training on coarse- and fine-grained feedback are feasible; for instance, here, we only consider \textit{point-wise} scores, rather than pairwise-based training.

We use the same model architecture for Stage 1 of all reward models. Models take the form of an MLP with two 256-dimension hidden layers and are trained for 100 epochs, with a learning rate of 1e-4. We use a batch size of 128. MLP training is implemented in jax. Linear aggregators are trained with class-balancing using the Logistic Regression scikit-learn model; all other sckit-learn defaults were used.

\subsection{Additional Details on Forms of Fine-Grained Feedback}

\subsubsection{Image Quality}
\label{granular-img-quality-details}

We query PaLI, a large-scale language-and-text model\citep{pali2023}, as to whether a given image meets a particular attribute. Specifically, we ask yes/no questions of the form: ``is the image [attribute]'' where attribute is $\in \{$blurry, distorted, visually compelling, captivating, funny, photorealistic, bright, disturbing, chaotic$\}$. We select these subset of attributes to span a range of axes along which one may consider eliciting feedback: positive / negative framing; relevant / irrelevant. We normalize the resulting scores as a softmax over the ``yes'' and ``no'' returned scores.

\begin{figure}[h]
\begin{center}
\includegraphics[width=0.45\textwidth]{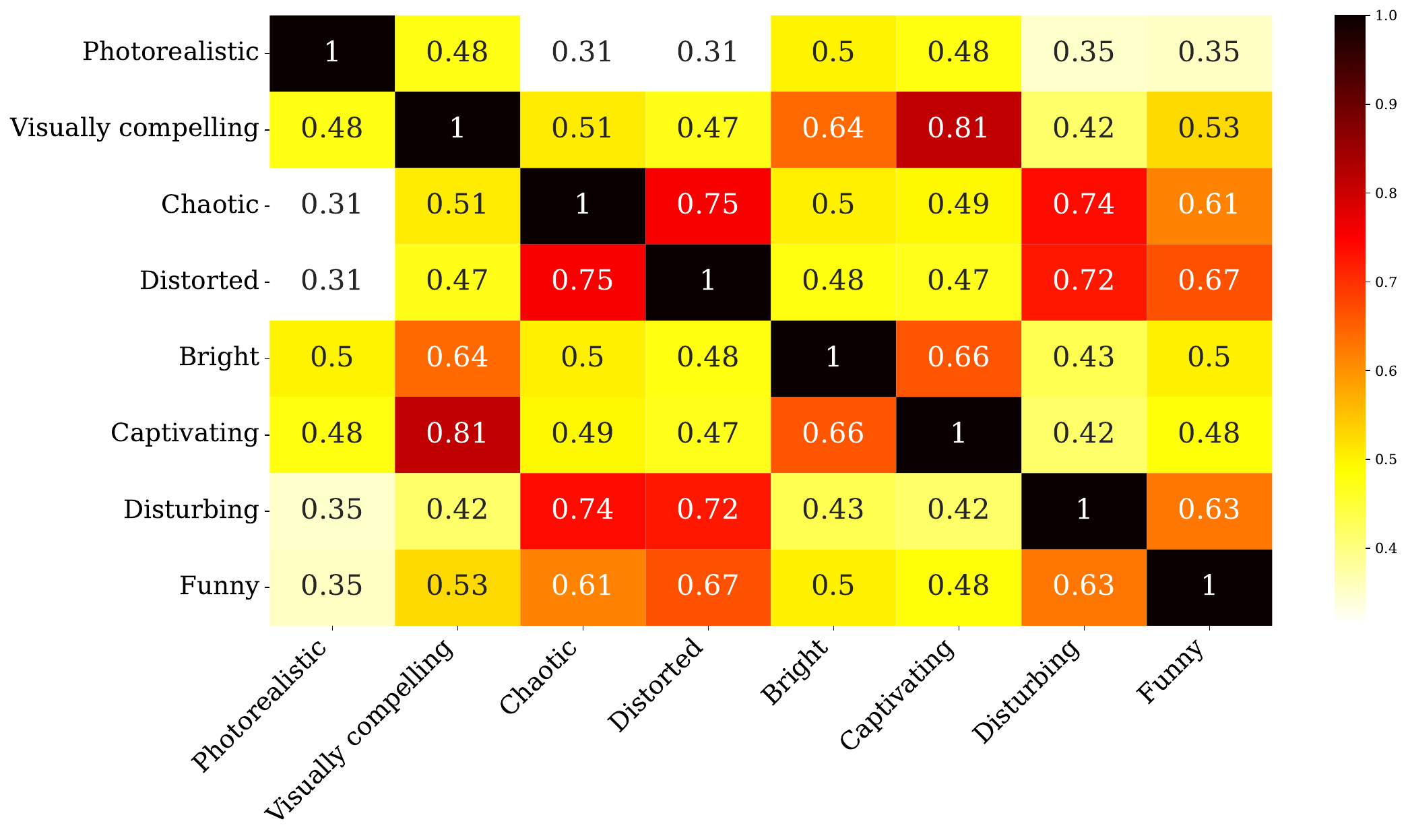}
\end{center}
\caption{Estimated similarity between PaLI scores for different attributes. We depict the proportion of images in the train set for which PaLI marks an image as having the same attribute (e.g., the cell blurry and malformed highlights that PaLI marks an image as blurry and malformed, or not blurry and not malformed for 71\% of the examples). Darker red means higher level of similarity in scores; yellow represents lower similarity.}
\label{fig:pali_attr_scores}
\end{figure}

\subsubsection{Prompt Alignment}
\label{granular-text-to-image-alignment-details}

We build on the $VQ^2$ method developed in \citep{yarom2023you} to measure the alignment between a prompt and the image. As discussed, $VQ^2$ takes as input an image and a prompt (e.g., ``green dog to the left of the river'') and generates a series of binary questions that the image ought to address if fit for the prompt (e.g., ``is there a dog?'', ``is the dog green?'', ``is the dog running along the left of the river?''). Each question is then assessed against the image, wherein the probability that the question can be answered as ``Yes'' is computed. The mean of the ``Yes'' probabilities forms the final $VQ^2$ score.

Here, we construct four scores to reflect different ways in which an image may be aligned to a prompt. An image may be aligned in its: 1) representations of objects / nouns (e.g., ``is there a dog?''), 2) attributes / adjectives (``is the dog green?''), 3) actions / verbs (``is the dog running?''), and 4) relations (``is the dog to the left of the river?''). We categorize each of the questions generated by $VQ^2$ into one of these categories using a custom semantic parser built from spaCy ~\citep{honnibal2020spacy} with hand-crafted rules to catch exceptions. Two authors from our author team manually inspected hundreds of the categorizations to affirm their quality -- while the parsing was generally sensible -- we note that it is not perfect and likely could be improved in future work. We then average the $VQ^2$ scores for all questions grouped in a category, which are then thresholded into a binary aligned/not aligned which we use as feedback. Image-prompt pairs for which $VQ^2$ does not generate a question for a particular class are binarized into the positive (i.e., aligned) class as we care more about cases which are mis-aligned along an attribute. We encourage future work to improve both the question generation, classification, and answer categorization.

\subsection{Additional Details on Data Processing}

We form preference judgements by aggregating over the contextually-annotated images from ~\citep{dutta2024understanding}. We apply simple averaging, where each annotation is weighted equally -- alternate weighting schemes could be worth exploring in the future, as well as a breakdown by the context. We split the data along the prompts, as there are four different images per prompt, each annotated with human scores.

The models we consider in this work involve binary classification; as such, we binarize all scores -- for the aggregate and fine-grained preference judgments. Thresholds are selected manually via a mix of attempting to class-balancing and manual inspection. Future work can explore more expansive threshold selection.

\subsection{Additional Details on Rejection Sampling}
\label{rejection-sampling-details} 

\subsection{Stimuli Generation}

We sample images from a generative text-to-image model similar to \citeauthor{rombach2022high}, trained on web-scale image data, using the prompts from the test set of  ~\citeauthor{dutta2024understanding}. To that end, our stimuli are slightly out-of-distribution (in-distribution prompts, out-of-distribution generated images).

\subsection{Reward Model Scoring and Selection}

We run two reward models (one coarse-, one fine-) over all generated prompt-image pairs. We apply our same embedding extraction pipeline and concatenate the text and image embeddings. We extract an aggregate reward score from each reward model. We select a subset of 194 text-image pairs where the reward models substantially differ in their preference judgements. 

\subsection{Human Study: SxS Evals}
We conducted a side-by-side evaluation of 194 pairs of images selected through the reward model scoring, where one image in the pair is scored highly by the fine-grained model and the other, scored highly by the coarse-grained model. The participants were asked to select an image that they preferred (general preference experiment) or asked to select and image that was ``more X'', where X is one of the features used for the fine-grained model (e.g., \textit{bright, funny}). Judgments for each feature were collected in separate tasks, leading to a total of nine tasks (eight fine-grained features and one general preference judgment task). The participants had the option to answer ``unsure''. We recruited three participants per image pair through an internal crowdsourcing platform. All of the questions and the sides for each question (left/right) were randomly shuffled.

\subsection{Visualizing the Learned Aggregators}

\begin{figure*}[h]
\begin{center}
\includegraphics[width=\textwidth]{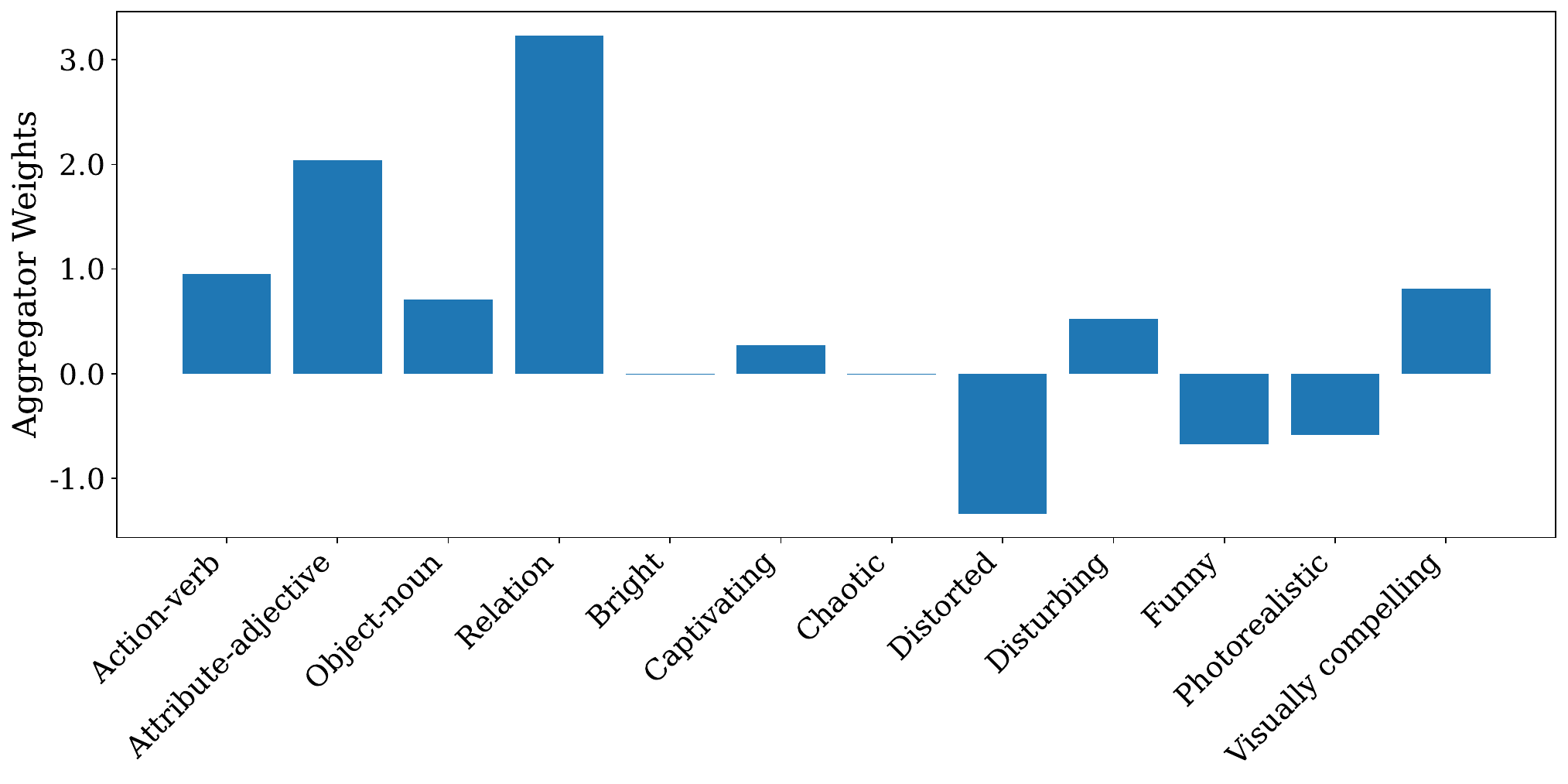}
\end{center}
\caption{Attribute weights learned for the prompt-aware setting; most weight is placed on attributes scoring prompt-image alignment.}
\label{fig:promptaware-importances}
\end{figure*}

\begin{figure*}[h]
\begin{center}
\includegraphics[width=\textwidth]{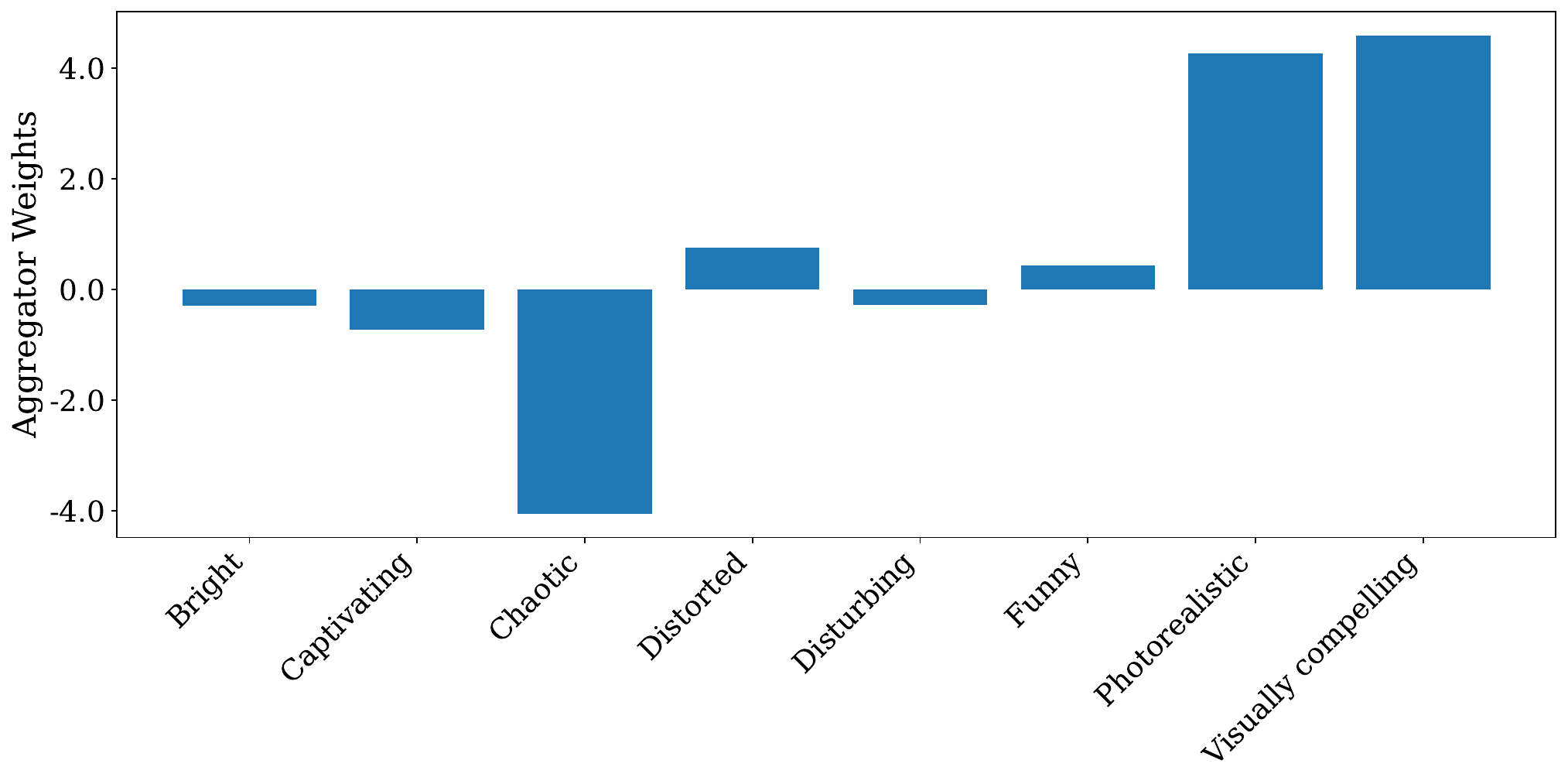}
\end{center}
\caption{Attribute weights learned for the decision tree setting. The aggregator appropriately learns the importance and direction of the attributes which make up the leaves of the tree.}
\label{fig:dectree-importances}
\end{figure*}

\begin{figure*}[h]
\begin{center}
\includegraphics[width=\textwidth]{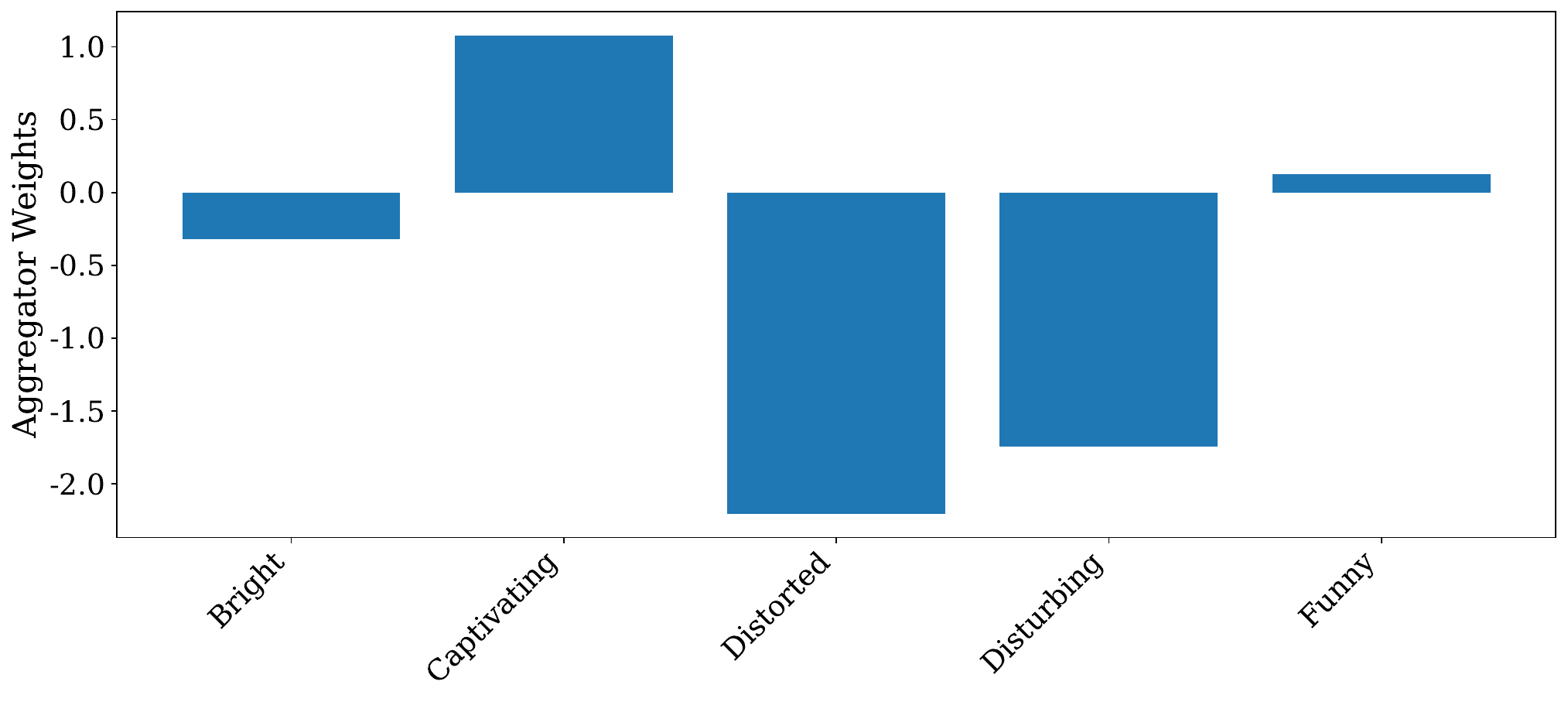}
\end{center}
\caption{Attribute weights learned for the decision tree setting. The aggregator learns to focus on attributes semantically related to those which form the decision tree (e.g., captivating versus visual compelling).} 
\label{fig:dectree-missing-leaves}
\end{figure*}

One of the advantages of the CBM structure of our reward model is that humans can inspect, and therefore audit, the attributes that are learned and most contribute to the final reward preference by inspecting the linear aggregator. We depict the linear aggregator weights for a sampling of the models in Experiment 1 (Figure \ref{fig:promptaware-importances}) and Experiment 2 (when the decision tree attributes were included in Figure \ref{fig:dectree-importances} and when missing \ref{fig:dectree-missing-leaves}).

\subsection{Samples Scored by Fine-Grained Models}
\label{model-samples}

\begin{figure*}[h]
\begin{center}
    \includegraphics[width=1.0\textwidth]{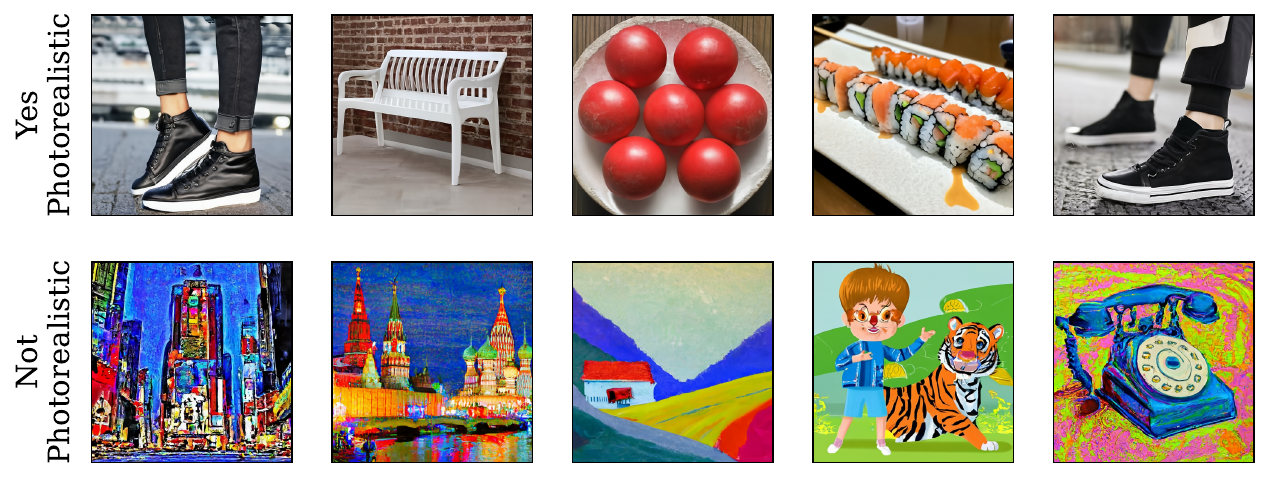}
    \includegraphics[width=1.0\textwidth]{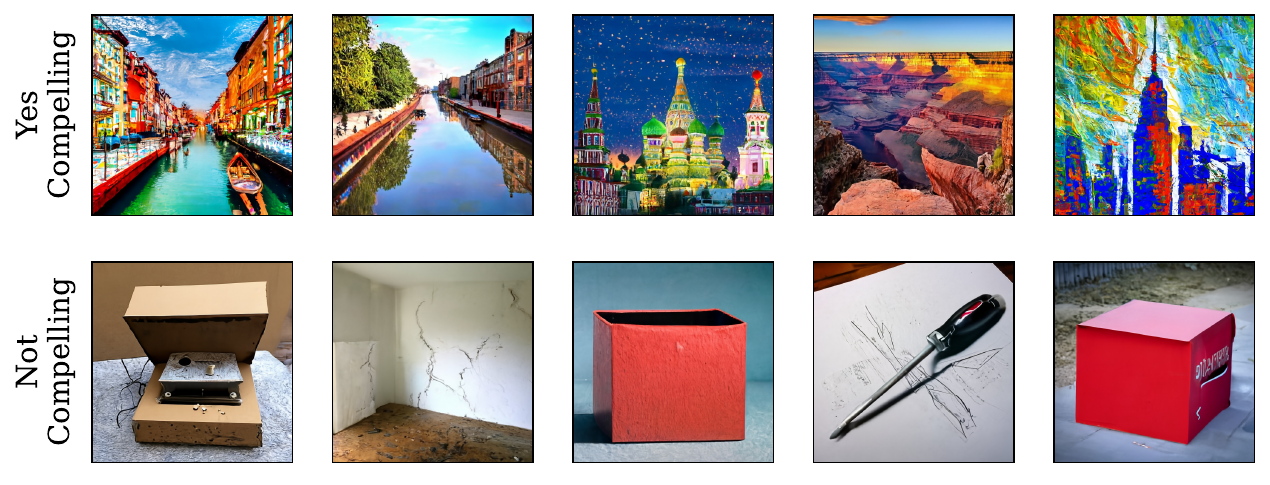}
    \includegraphics[width=1.0\textwidth]{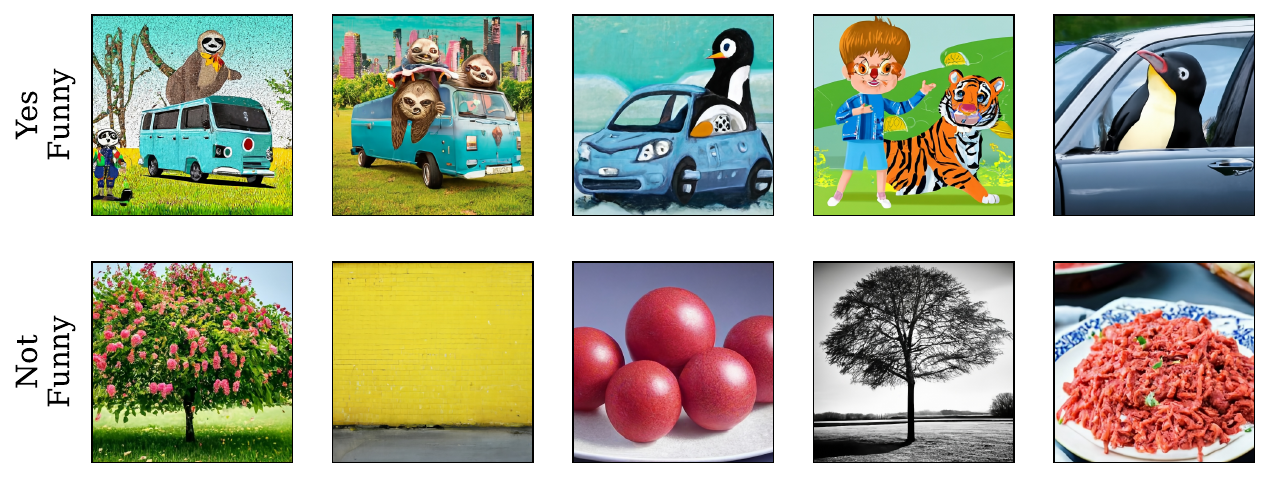}
    \caption{Images scored by the trained fine-grained reward model. Images in the top row are those which are rated high on an attribute (noted on the y axis), and in the bottom row, rated low by the reward model in terms of that attribute. Note, ``compelling'' here is ``visually compelling''.}
    \label{fig:viz-reward-models}
\end{center}
    
\end{figure*}

We include some images from our data pool which were scored as good or bad along a sampling of attributes in Figure \ref{fig:viz-reward-models}. 

\end{document}